\documentclass[sigconf,screen]{acmart}

\AtBeginDocument{%
  \providecommand\BibTeX{{%
    \normalfont B\kern-0.5em{\scshape i\kern-0.25em b}\kern-0.8em\TeX}}}

\renewcommand\footnotetextcopyrightpermission[1]{} % removes 

\usepackage{multirow}
\usepackage{enumitem}
\usepackage{bm}

\usepackage{fancyhdr}
\begin{document}

\title{Making Your Dreams A Reality: Decoding  the Dreams into a Coherent Video Story from  fMRI Signals\footnotemark[2]}

\author{Yanwei Fu\footnotemark[1], Jianxiong Gao, Baofeng Yang, Jianfeng Feng}

\begin{abstract}

This paper studies the brave new idea for Multimedia community, and proposes a novel framework to convert dreams into coherent video narratives using fMRI data. Essentially, dreams have intrigued humanity for centuries, offering glimpses into our subconscious minds. Recent advancements in brain imaging, particularly functional magnetic resonance imaging (fMRI), have provided new ways to explore the neural basis of dreaming. By combining subjective dream experiences with objective neurophysiological data, we aim to understand the visual aspects of dreams and create complete video narratives. Our process involves three main steps: reconstructing visual perception, decoding dream imagery, and integrating dream stories. Using innovative techniques in fMRI analysis and language modeling, we seek to push the boundaries of dream research and gain deeper insights into visual experiences during sleep. This technical report introduces a novel approach to visually decoding dreams using fMRI signals and weaving dream visuals into narratives using language models. We gather a dataset of dreams along with descriptions to assess the effectiveness of our framework.
  
\end{abstract}

% \keywords{Neural Decoding, Functional MRI, Diffusion Model, Sleep, Dream}

\begin{teaserfigure}
  \includegraphics[width=\textwidth]{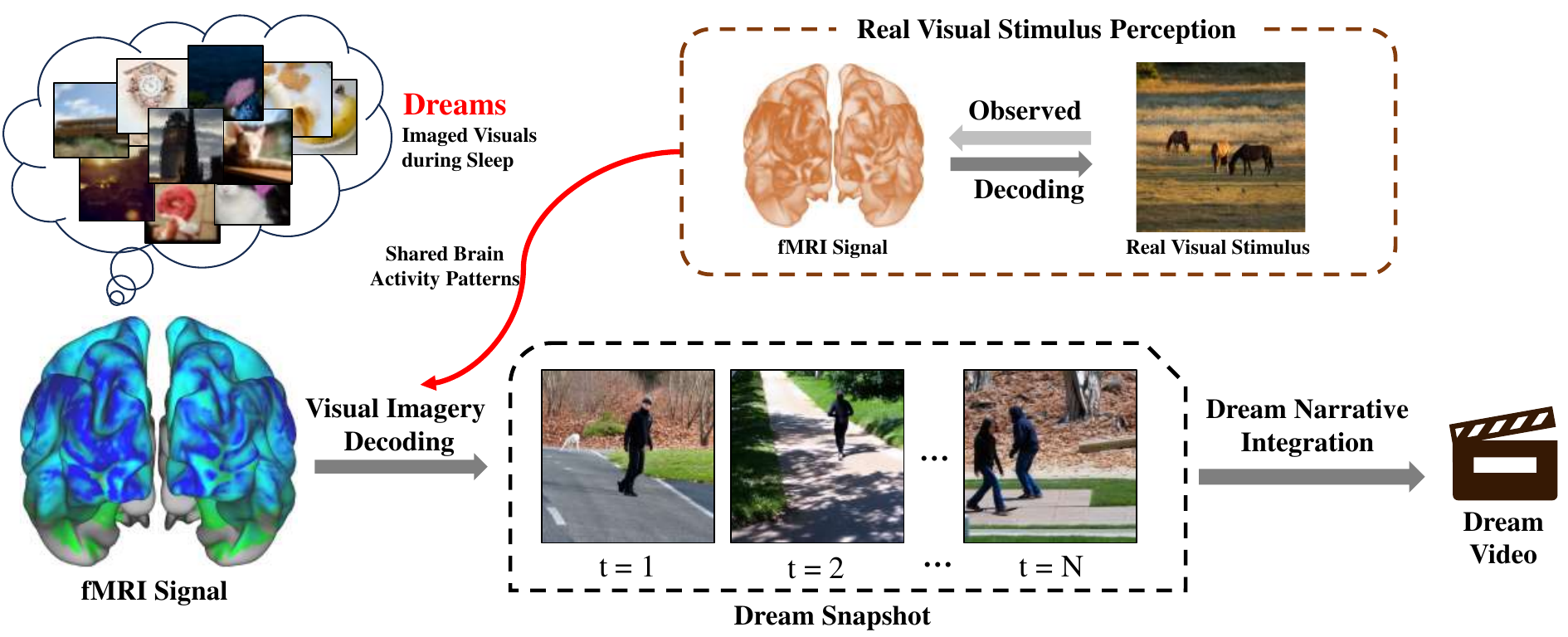}
  \caption{Our Dream decoding process blends fMRI decoding with tasks related to real and imagined visuals, seamlessly turning disjointed dream images into complete stories.
  }
  \label{fig:teaser}
  \vskip 0.05in
\end{teaserfigure}

\maketitle

\renewcommand{\thefootnote}{\fnsymbol{footnote}} 
\footnotetext[1]{Corresponding authors. Email: yanweifu@fudan.edu.cn}
\footnotetext[2]{Work in progress}

\section{Introduction}

Dreams have always captured the imagination of people, from scientists and philosophers to artists. Although dreams give us a peek into the hidden parts of our minds, revealing the intricate ways we think and feel, they are still shrouded in mystery.  
For a long time, the fleeting and deeply personal nature of dreams has made them hard to study. However, thanks to new brain imaging tools, especially functional magnetic resonance imaging (fMRI), we are starting to understand what happens in the brain when we dream. Using fMRI, researchers can now see which parts of the brain are active during sleep and relate these patterns to the stories and emotions people describe after they wake up.
We consider the task of turning dreams into coherent video stories a very brave new idea in our Multimedia Community. It offers the unique opportunity to transform your dreams into reality using a completely new type of input, the fMRI brain scans, and helps us understand some basic capabilities of humans.

The main goal of our task is to use fMRI to transform the visual elements of dreams into detailed videos, going beyond basic categorization to fill gaps in our current understanding of dream content. 
Analyzing the possibilities and challenges of this task, we recognize that while people can describe their dreams, fMRI reveals the specific brain areas involved in creating and processing these dreams. By combining personal accounts with brain imaging, researchers can start identifying the brain's "signatures" for different aspects of dreaming, such as memory, emotion, and sensory experiences during sleep. Through analyzing brain activity during dreams and employing sophisticated techniques to decode these signals, we achieve a high-fidelity reconstruction of visual experiences within dreams, surpassing the limitations of semantic categorization. However,  the primary challenge in interpreting dreams is \textit{bridging the gap between subjective experiences and objective neurophysiological data}.

Furthermore, dreams are not merely isolated images but are composed of cohesive, narrative sequences~\cite{hobson2000dreaming}. The conventional method~\cite{horikawa2013neural} of decoding individual dream images falls short in capturing the continuous brain activity that generates dream experiences. Therefore, we propose designing a novel toolkit to automatically put together different media elements decoded based on what the user dreams, and create
a coherent video story reflecting the dreams. Critically, our novel toolkit also presents the textual descriptions by leveraging the Large Language Models (LLMs)~\cite{zheng2024intelligent} to interpret the decoded images and integrate them into comprehensive dream experiences.

As the brave new idea for this novel applications, there is no previous work or research that conducts this task. We thus propose a novel pipeline that, for the first time, enables neurally decoding dreams. Particularly, as shown in Fig.~\ref{fig:teaser}, our approach involves several important steps: modeling how dreams appear visually in the brain, combining fMRI decoding with tasks involving both real and imagined visuals, and turning disjointed dream images into complete narratives.

We give the detailed process of decoding dreams from fMRI scans into three main steps. First, in the Reconstructing Visual Stimulus Perception stage, we decode brain activity linked to seeing real objects or scenes by analyzing fMRI data collected when subjects are awake and viewing visual stimuli, aiming to recreate these images based on brain processes. Next, in the Decoding Dream Visual Imagery stage, we use brain activity patterns from both awake visual experiences and dream visuals to decode images from the dream, analyzing sleep-state fMRI data to capture dream visuals as they occur at different times. Finally, in the Integrating Dream Narratives stage, we stitch together the decoded images into a full, flowing dream story using advanced language models, aiming to create a cohesive narrative that captures the entire dream experience.

This is a `brand new' novel idea of making your dreams a reality. Our central contribution is to make this incredible task work. We summarize the key technical novelties that are important to our tasks. \\
(1) We make a first try of introducing a novel method using fMRI signals to visually decode the dreams, and enabling dreams into a coherent video story. \\
(2) We propose bridging the gap between real and dream visual experiences, providing deeper insights into dream generation; \\
(3) We present integrating dream visualizations into complete visual narratives using LLMs, which helps us better understand what dreams mean.

We collect a dataset of dreams along with detailed descriptions to evaluate how well our framework works. Usually, it's quite challenging for participants to provide precise descriptions of their dreams. As a result, the gathered descriptions mostly serve as general indicators to gauge the effectiveness of our framework.

\section{Related Work}
\label{sec:related}

\subsection{Dream Visual Decoding}

Dreams often present themselves as disjointed and fantastical scenes, plots, and feelings.
Early research suggested that similarities in these phenomena stem from a shared neural substrate between wakefulness and sleep states. Several studies analyzed brain region activations across different states to investigate the neural commonalities and differences \cite{maquet1996functional, braun1997regional, hong2009fmri}.
Moreover, Northoff et al.~\citep{northoff2023topographic} introduced a specific spatiotemporal model of dreams to bridge the gap between neural fluctuations and mental experiences. However, these studies have not precisely determined how specific visual content is represented in brain activity. Horikawa et al.~\cite{horikawa2013neural} pioneered the use of functional magnetic resonance imaging (fMRI) and machine learning analysis to decode spontaneous brain activity during sleep to identify visual content in dreams. Subsequently, hierarchical visual feature decoding techniques were employed to determine brain decoding of objects observed or imagined by individuals \cite{horikawa2017generic}, and more accurately, dream object decoding \cite{horikawa2017hierarchical}.

However, these methods of visual decoding only achieve semantic category-level classification, making it challenging to visualize the real visual experiences of dreams. Therefore, this paper aims to fill this research gap by utilizing state-of-the-art fMRI-to-image reconstruction techniques to achieve precise reconstruction of the fine-grained visual experiences in dreams.

\subsection{Visual Stimulus Decoding}

Compared to Dream Visual Decoding, decoding fMRI signals under real visual stimuli has been studied more extensively. Visual Stimulus Decoding relies on observed images and their corresponding fMRI responses to achieve fMRI-to-image reconstruction. Early Visual Stimulus Decoding work primarily focused on high-level semantic information of images \cite{cox2003functional, haxby2001distributed, thirion2006inverse, kay2008identifying}, or specific categories of image tasks such as face recognition. The development of generative models like GANs and VAEs subsequently led to a trend in decoding natural images from fMRI \cite{shen2019deep, mozafari2020reconstructing,ren2021reconstructing, gu2022neurogen}. However, the images decoded by these methods still suffer from blurriness and distortions. Recently, benefiting from the powerful generative capabilities of Diffusion Models \cite{rombach2022high}, researchers have used fMRI signals as conditions for Diffusion Models to obtain high-quality visual reconstruction results \cite{chen2023seeing,ozcelik2023natural, scotti2023reconstructing,zeng2023controllable,ferrante2023brain, fang2024alleviating}. Chen et al. \cite{chen2023seeing} decomposed fMRI-to-image reconstruction into fMRI embedding learning based on MAE and conditional attention layers based on fine-tuning LDM, improving the semantic relevance and visual quality of reconstruction results. Ozcelik et al. \cite{ozcelik2023natural} and Scotti et al. \cite{scotti2023reconstructing} proposed learning mappings from fMRI signals to CLIP text and image features, VAE latent codes respectively, leveraging high-level CLIP text and image features and low-level intermediate images to improve the consistency between reconstruction results and original visual inputs at the low level. However, these works can only be trained on individual subjects and cannot learn shared knowledge among multiple individuals. Qian et al. \cite{qian2023fmri} utilized 40k subjects from the UK Biobank dataset (UKB) \cite{miller2016multimodal} to train a large-scale fMRI representation model, which can encode multi-individual brain signals into a unified large-scale latent space. NeuroPictor \cite{huo2024neuropictor} then adopts fMRI-to-image cross-subject pre-training based on this unified representation to learn shared visual perception among multiple individuals and proposes using coupled low-level manipulation and high-level guiding networks to recover original images.

Inspired by the multi-subject pre-training fMRI-to-image methods in Visual Stimulus Decoding, we aim to transfer the knowledge learned from Visual Stimulus Decoding, involving different individuals and shared brain activity patterns under real and dreamlike stimuli, to Dream Visual Decoding.

%-------------------------------------------------------------------------
\section{Problem setups}

Dream decoding involves analyzing fMRI data recorded from individuals during sleep to extract their virtual experiences, termed dreams.
In this endeavor, individuals are subjected to fMRI scans while they are in a state of sleep, resulting in the continuous recording of brain activity patterns throughout the sleep cycle. These recorded fMRI data form a spatiotemporal sequence denoted as $\{F_i\} (i=1, \cdots, n)$ where $n$ is the recorded fMRI temporal length, representing the dynamic neural activity occurring during sleep. 

The objective of dream decoding is to extract from this complex spatiotemporal sequence the virtual visual experiences, commonly referred to as dreams\footnote{Please note that we specifically refer to "dreams" to make it easier for our Multimedia community to understand the new task. However, it is important to acknowledge that in neurology, there are precise definitions for different stages of sleep, and only certain stages are categorized as dreams. For detailed definitions, please refer to \cite{foulkes1962dream}.} that individuals undergo during sleep. However, dreams manifest as intricate and often surreal amalgamations of images, scenarios, and emotions, which are subjectively experienced by individuals during sleep. To achieve this, two fundamental aspects need consideration: (1) Virtual Visual Frames Generation: At specific moments during sleep, individuals generate virtual visual frames encapsulating their dream experiences. These frames represent snapshots of the dream content perceived by the individual at distinct time points during the sleep cycle. (2) Visual Story Information: These individual virtual visual frames are not isolated entities but rather pieces of a larger narrative. When integrated, they form a cohesive visual story that conveys the progression and thematic elements of the dream experience. 

In summary, we need to associate the brain activity patterns at each time point with the corresponding virtual visual frames, yielding the $i$-th snapshot image $I_i$ corresponding to ${F_i}$. Subsequently, by dynamically assembling these snapshot images, we can construct a complete visual narrative of the dream, denoted as $V$.

%-------------------------------------------------------------------------
\section{Data Collection and preprocessing}

\begin{figure}[th]
  \centering
  \includegraphics[width=\linewidth]{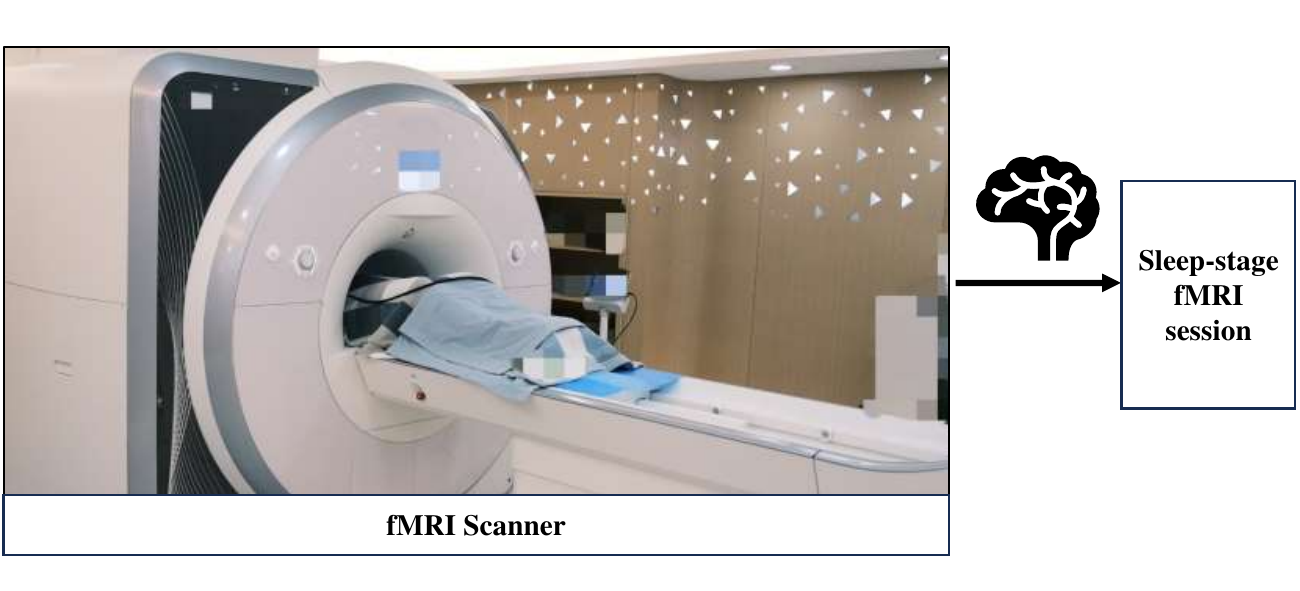}
  \caption{We use a 3 Tesla (3T) MRI scanner for data collection. Participants are positioned within the MRI scanner. }
  \label{fig:device}
  \vskip -0.05in
\end{figure}

During the fMRI sessions, participants are comfortably positioned within the MRI scanner, ensuring optimal alignment for the precise recording of their neural activity during the sleep cycle. Following~\cite{gao2023mind,gao2024fmri}, the data acquisition was conducted using a 3 Tesla (3T) MRI scanner and a 32-channel RF head coil with a temporal resolution of 1.25Hz.
To facilitate an in-depth exploration of sleep phenomena, volunteers were afforded a duration of sleep ranging from 40 to 70 minutes within the controlled environment of the MRI scanner. This extended period allowed for the exploration of various sleep stages characterized by distinct patterns of neural activity and physiological changes.
Throughout the data collection process, we employed monitoring devices to obtain participants' head movements and eyelid states, assisting us in determining whether participants remained in a stable sleep state throughout the recording session.

For preprocessing, we followed a processing pipeline~\cite{fmriprep1,fmriprep2} similar to that described in~\cite{qian2023fmri,gao2023mind} to map the neural activity of different participants onto a common surface map. This process facilitates the dissemination of knowledge across datasets, which is crucial for inter-individual understanding. Specifically, we convert the fMRI time series to fsLR32K surface space using Connectome Workbench, followed by z-scoring values across each session and rendering them into 2D images. 
This process yields brain surface images resembling a "butterfly" representing cortical activation. Finally, the early and higher visual cortical (VC) regions, including ``V1, V2, V3, V3A, V3B, V3CD, V4, LO1, LO2, LO3, PIT, V4t, V6, V6A, V7, V8, PH, FFC, IP0, MT, MST, FST, VVC, VMV1, VMV2, VMV3, PHA1, PHA2, PHA3'', are selected for further analysis. 

%-------------------------------------------------------------------------
\begin{figure*}[t]
  \centering
  \includegraphics[width=\linewidth]{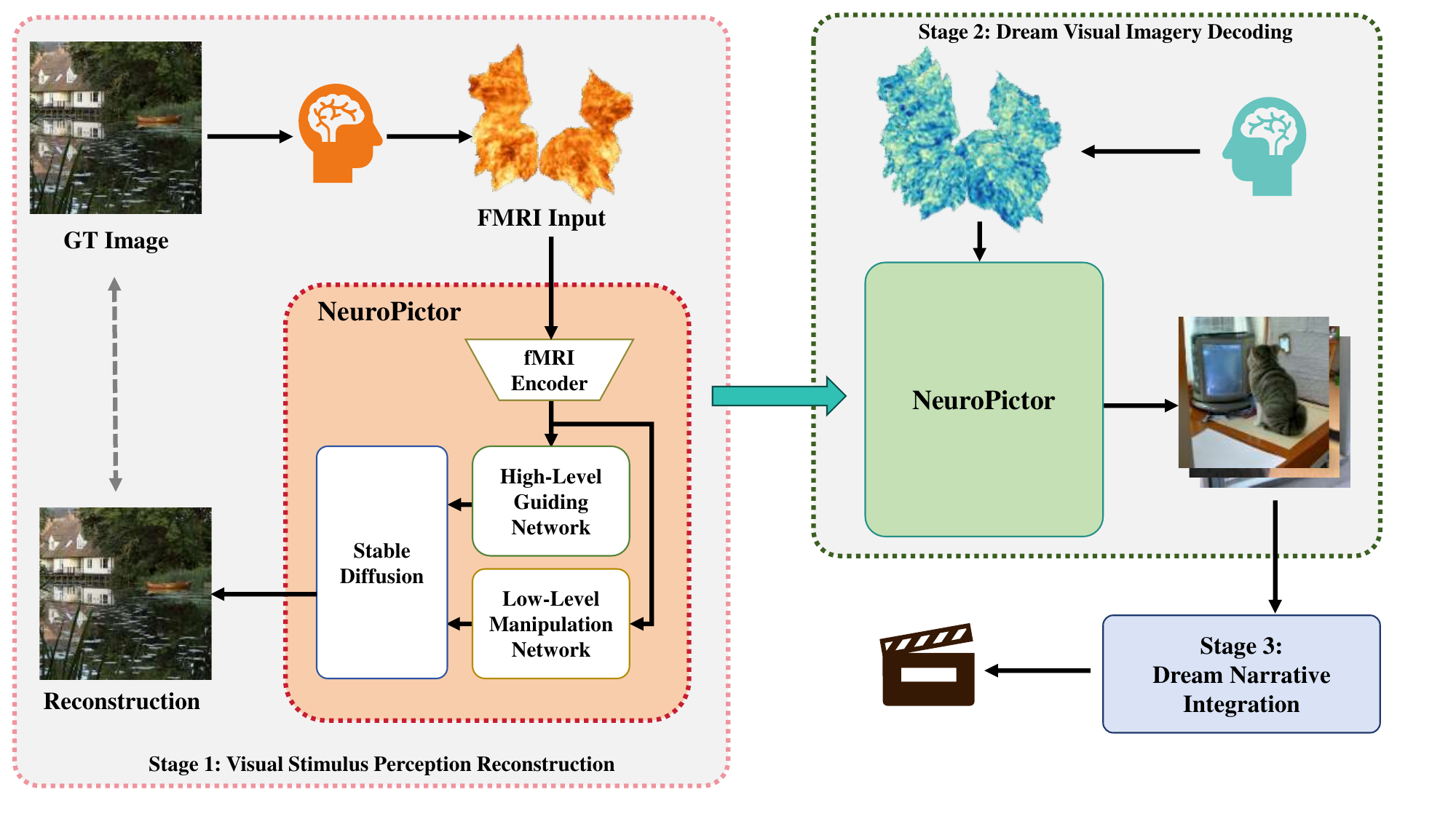}
  \caption{
  The fMRI dream decoding process is divided into three stages:
    i) Visual Stimulus Perception Reconstruction: Decoding brain activity patterns associated with real visual stimuli perception.
    ii) Dream Visual Imagery Decoding: Shared brain activity patterns between real visual stimuli and dream-induced visual experiences aid in decoding snapshots of dream content from fMRI data.
    iii) Dream Narrative Integration: By leveraging LLM, we synthesize fragmented dream visualizations into cohesive narratives, providing a complete interpretation of the dream experience.}
  \label{fig:pipeline}
\end{figure*}

\section{Method}

\noindent
{\bf Overview.}\quad
We divide fMRI dream decoding into three stages (illustrated in Figure \ref{fig:pipeline}): Visual Stimulus Perception Reconstruction, Dream Visual Imagery Decoding and Dream Narrative Integration. 

\noindent\textbf{i)} Visual Stimulus Perception Reconstruction: This stage involves decoding brain activity patterns associated with the perception of real visual stimuli. By analyzing fMRI data collected during wakefulness when subjects are presented with visual stimuli, we aim to reconstruct these stimuli using the neural representations.

\noindent\textbf{ii)} Dream Visual Imagery Decoding: In this stage, we leverage shared brain activity patterns across both real visual stimuli and dream-induced visual experiences to decode snapshots of the dream content. By analyzing fMRI data collected during the dream state, our aim is to decode the virtual visual frames representing the dream experiences perceived by individuals at distinct time points during the sleep cycle.

\noindent\textbf{iii)} Dream Narrative Integration: This final stage involves integrating the decoded single-frame dream visualizations into a comprehensive narrative of the entire dream sequence. Leveraging Large Language Models (LLM), we aim to synthesize fragmented dream visualizations into cohesive narratives, providing a complete interpretation of the dream experience.

%-------------------------------------------------------------------------
\begin{figure*}
  \centering
  \includegraphics[width=\linewidth]{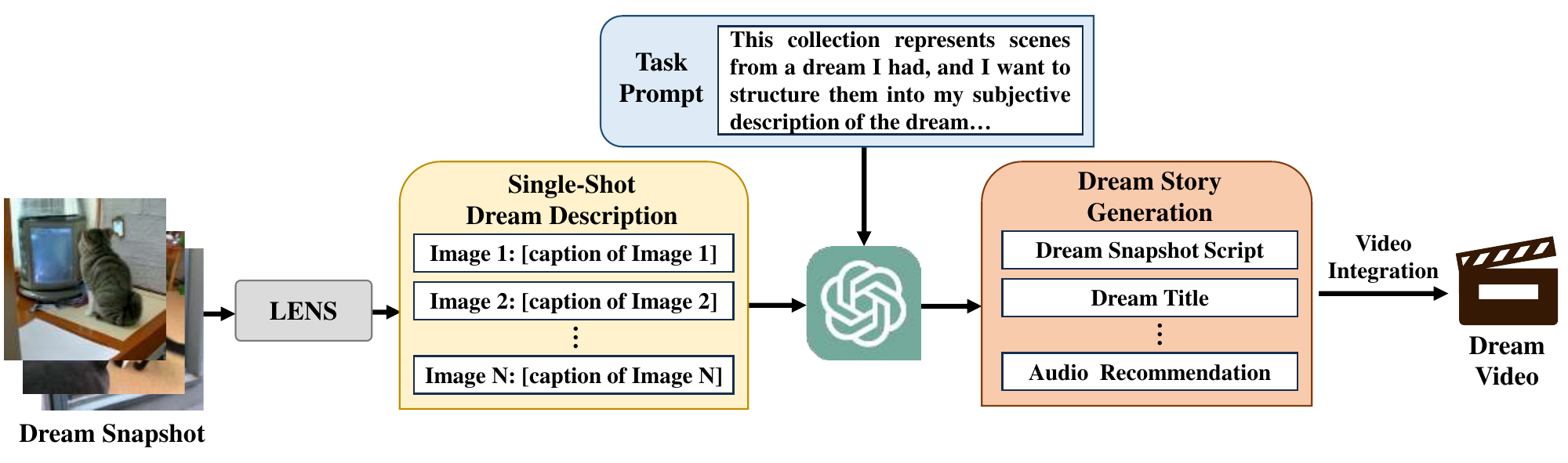}
  \caption{Pipeline of Dream Narrative Integration. This process includes three steps: Single-Shot Dream Description, Dream Story Composition, and Video Integration.}
  \label{fig:stage3}
\end{figure*}

%-------------------------------------------------------------------------
\subsection{Visual Stimulus Perception Reconstruction}
\label{sec:stage1}

There are two key challenges in directly using dream data for model training: data scarcity and the lack of real dream visual images. On one hand, collecting dream data is constrained by the fact that it can only be obtained during participants' sleep states, and the availability of effective data relies on the number of dreams recalled by participants. Dreams are spontaneously generated by the brain during sleep stages, and even with prolonged sleep, the number of accurately recallable dreams is limited. On the other hand, in terms of interpreting dream content, dream reports provide us with captions of the dreams, but these are limited to higher-level semantics. While such semantic-level labels can be used for training classification tasks, they are insufficient for reconstructing visual images from dreams, making supervised training challenging.

Therefore, to address these issues, we propose to introduce an fMRI-to-image task under real visual stimuli. In the Visual Stimulus Perception Reconstruction task, subjects observe specific images while brain activity is recorded, and then the task involves decoding real visual stimuli from fMRI signals. This dataset of fMRI-image pairs is significantly larger in scale than dream datasets, thus addressing the issue of original data scarcity. Moreover, inspired by research \cite{horikawa2013neural} suggesting that specific visual experiences during sleep are represented by brain activity patterns shared by stimulus perception, we attempt to utilize shared brain activity patterns across both real visual stimuli and dream-induced visual experiences to decode snapshots of dream content. Specifically, we first learn to reconstruct real visual stimuli from fMRI signals, and then transfer the learned brain activity patterns associated with real visual stimuli to dreamlike visual perceptions.

We base our study on the Natural Scenes Dataset (NSD)~\cite{allen2022massive} for learning the Visual Stimulus Perception Reconstruction task. NSD comprises fMRI-image data from 8 individuals, with the original image data consisting of approximately 65k images from the MS-COCO~\cite{lin2014microsoft} dataset. This large-scale dataset provides a basis for learning cross-individual, generalizable fMRI-to-image reconstruction. We follow NeuroPictor \cite{huo2024neuropictor} for the reconstruction process as it allows for training a unified model across multiple subjects. Specifically, we utilize the fMRI surface map $F$ as input, initially processed through a transformer encoder to obtain a unified fMRI representation across different persons:
\begin{equation}
    {F}^r = \operatorname{E}(F),
\end{equation}
where $\operatorname{E}$ denotes the fMRI encider. 
Then, we employ a diffusion generative model integrated from a High-Level Guiding Network and Low-Level Manipulation Network to guide semantic and low-level details' generation as follows:
\begin{equation}
I^{real} = \operatorname{G}(F^r),
\end{equation}
where $I^{real}$ is the reconstructed image, and $\operatorname{G}$ is the integrated generative model. We follow the training objectives in \cite{huo2024neuropictor} on the NSD dataset of eight individuals, facilitating subsequent knowledge transfer between different individuals and between real visual responses and dream-induced visual responses.

%-------------------------------------------------------------------------
\subsection{Dream Visual Imagery Decoding}
\label{sec:stage2}

Building upon the Visual Stimulus Perception Reconstruction model obtained from the previous section, we aim to accomplish Dream Visual Imagery Decoding by leveraging shared brain activation patterns between real and dream experiences. Benefiting from the large scale and image diversity of the NSD dataset, our trained generative model can cover different individuals and a wide range of images. Considering previous research suggesting shared visual cortex activation responses between real and dream experiences, we directly transfer the trained Visual Stimulus Perception Reconstruction model to the Dream Visual Imagery Decoding task. For the fMRI sequences collected during sleep stages, we first average them according to a window size to align with the fMRI acquisition process under real visual stimuli, where participants need to observe images for 3-4 seconds. Then, we perform zero-shot decoding on each discrete timestamp of the fMRI to obtain the dream visual decoding image corresponding to the i-th moment:
\begin{equation}
I^{dream}_i = \operatorname{G}(\operatorname{E}(F_i)),
\end{equation}
where $F_i$ is the original fMRI signal in the i-th moment and $I^{dream}_i$ is the corresponding dream visual image decoded by $F_i$.

In contrast to previous approaches in dream decoding, which relied on dream report labels to provide only coarse-level supervision for learning to decode the types of objects present in dreams, our method directly decodes visual images from dream experiences. This direct decoding approach allows us to visualize the content of dreams directly rather than conducting rough category-level classification.

%-------------------------------------------------------------------------
\subsection{Dream Narrative Integration}
\label{sec:stage3}

Reconstructing dream visuals frame by frame can only discretely represent the dream scenes generated by participants at specific time points. However, the dream experiences formed by individuals during the sleep stage unfold as a continuous narrative.

We turn to the assistance of large-scale language models to accomplish the integration of dream stories, by employing these models to bridge the fragments into a cohesive narrative. This endeavor can be defined as follows: given discrete dream visual reconstruction results $\{I^{dream}_i\} (i=1,\cdots,N)$, we aim to dynamically compose these individual shots into a unified story. Inspired by the Intelligent Director Framework proposed in \cite{zheng2024intelligent} for automated video synthesis using ChatGPT \cite{brown2020language}, we adjust this pipeline to fit the task of Dream Narrative Integration. This process can be divided into three steps: Single-Shot Dream Description, Dream Story Composition, and Video Integration.

In the Single-Shot Dream Description phase, we first employ the Image-Text QA model LENS~\cite{berrios2023language} to generate textual descriptions for each individual shot's dream visual reconstruction $I^{dream}_i$. This provides a detailed description of each dream shot $I^{dream}_i$. Subsequently, we organize these descriptions into a sequentially arranged caption prompt using the following template:

\begin{quote} %\vspace{2pt}
\small
% \textbf{caption template}: 
\vspace{-3.5pt}
\begin{equation*}
\begin{aligned}
\text{`` }&\text{Image 1: [caption of Image 1]} \\
&\text{Image 2: [caption of Image 2]} \\
\vspace{-4pt}
&~~~~~~~~~~~~~~~~~~~~~~~~~~~~~~~~~~~~~~~~~~~~~\vdots \\
\vspace{-4pt}
&\text{Image N: [caption of Image N] ''.}
\end{aligned}
\end{equation*}
\vspace{-3.5pt}
\end{quote} %
\vspace{-3pt}

During the Dream Story Composition phase, we utilize ChatGPT to generate a logically coherent narrative with a dream-like storytelling style. We tailor a question prompt to transform our dream narrative integration task into ChatGPT-based script generation based on the captions of individual dream shots. The prompt for task description is defined as follows:

\begin{quote} %\vspace{2pt}
``
{\small
I have a collection of photos and videos, with a fixed order. I need your help to organize these materials according to their input sequence. This collection represents scenes from a dream I had, and I want to structure them into my subjective description of the dream based on their captions. Additionally, I require a smoothly written script that connects these images into a cohesive narrative.

I need you to do two things:

(1) Provide a subjective description of my dream from my perspective based on the captions of these images and videos, keeping the order fixed according to the input sequence.

(2) Write a script according to the input material sequence. The script should be concise, fluent, vivid, and the transitions between different materials should be natural. }
''
\end{quote} %\vspace{3pt}

In addition to the task description prompt, we also customize a standardized output prompt. To achieve a harmonious blend of audio and visual elements in the dream story video, we request ChatGPT to generate a script for each dream shot, including dream titles, concluding remarks, and a recommended audio track. The complete prompt template is available in the supplementary materials. Thus, we utilize ChatGPT to integrate logically coherent dream narrative textual material.

Finally, based on the dream narrative text generated by ChatGPT, we proceed with video integration. Specifically, following the sequence of dream shots, we insert corresponding scripts for each shot and combine them with dream titles, concluding remarks, and recommended audio tracks to create a comprehensive video. This final video encompasses dream imagery, narrative text, and suitable music, thereby reflecting a complete dream experience during the sleep stage.

\section{Experiments}

\begin{figure*}
  \centering
  \includegraphics[width=0.65\linewidth]{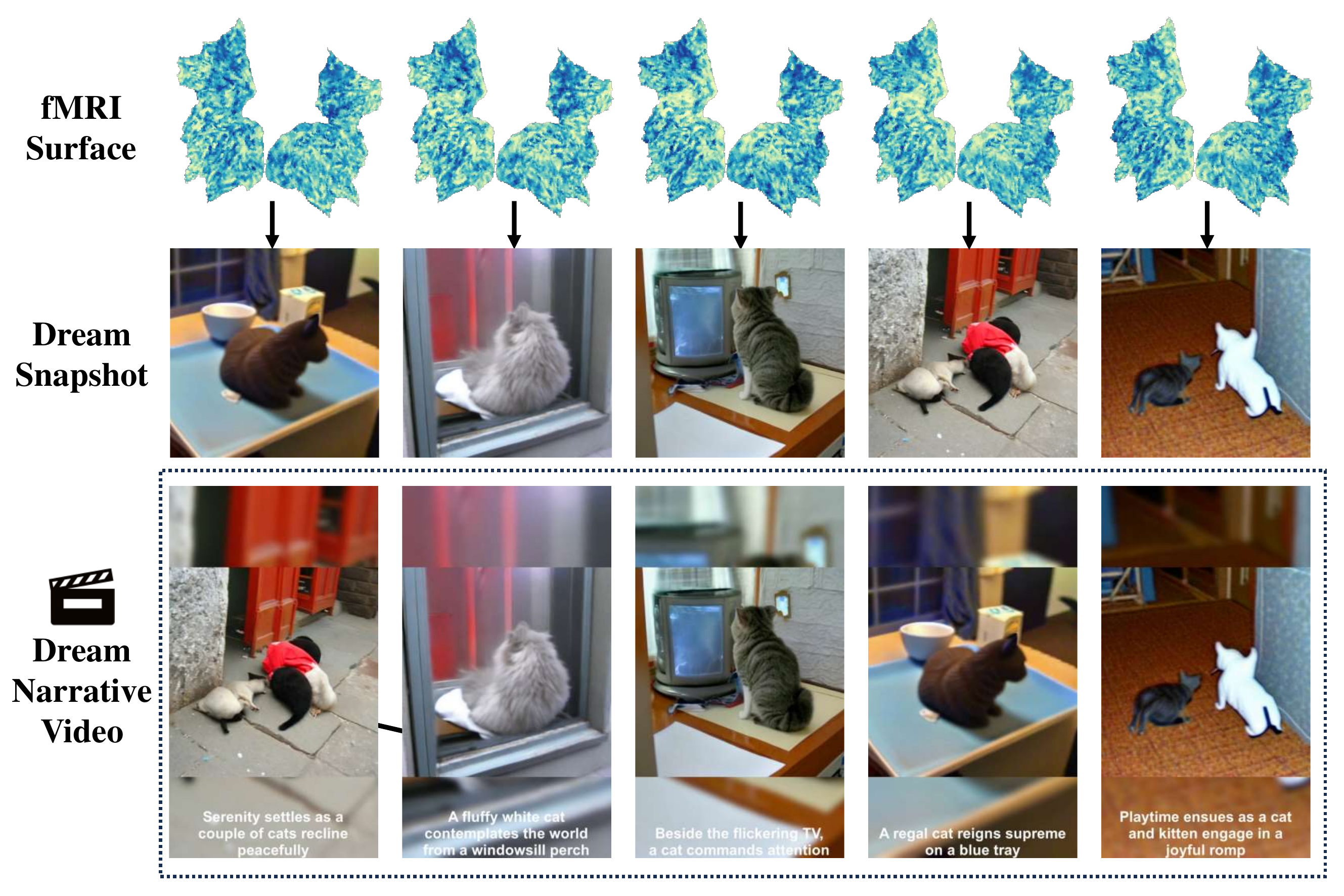}
  \caption{Visualization of Dream Narrative Video "some cat". }
  \label{fig:cat}
\end{figure*}

\begin{figure*}[t]
  \centering
  \includegraphics[width=0.6\linewidth]{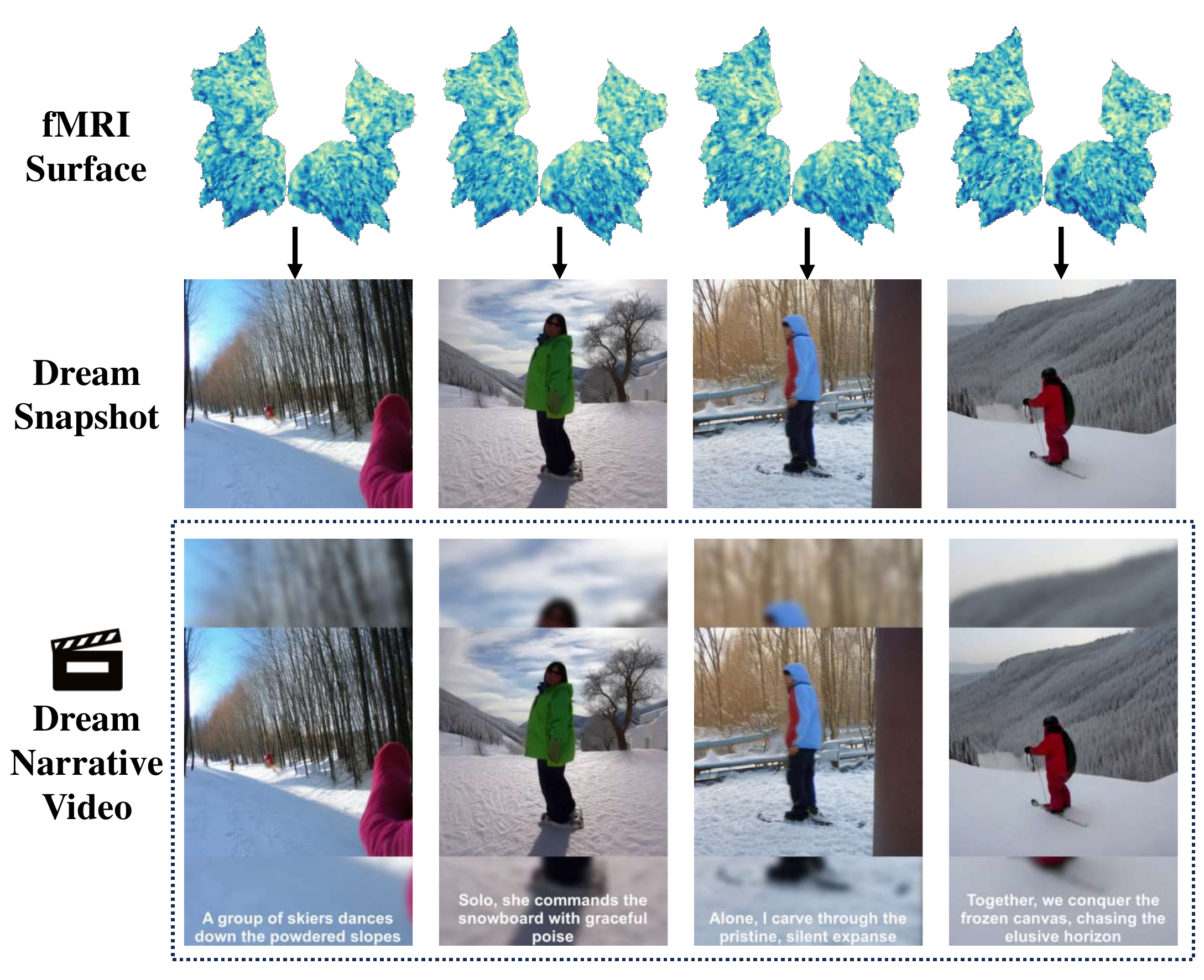}
  \caption{Visualization of Dream Narrative Video "skiing with a snowboard". }
  \label{fig:ski}
\end{figure*}

\begin{figure*}[t]
  \centering
  \includegraphics[width=0.9\linewidth]{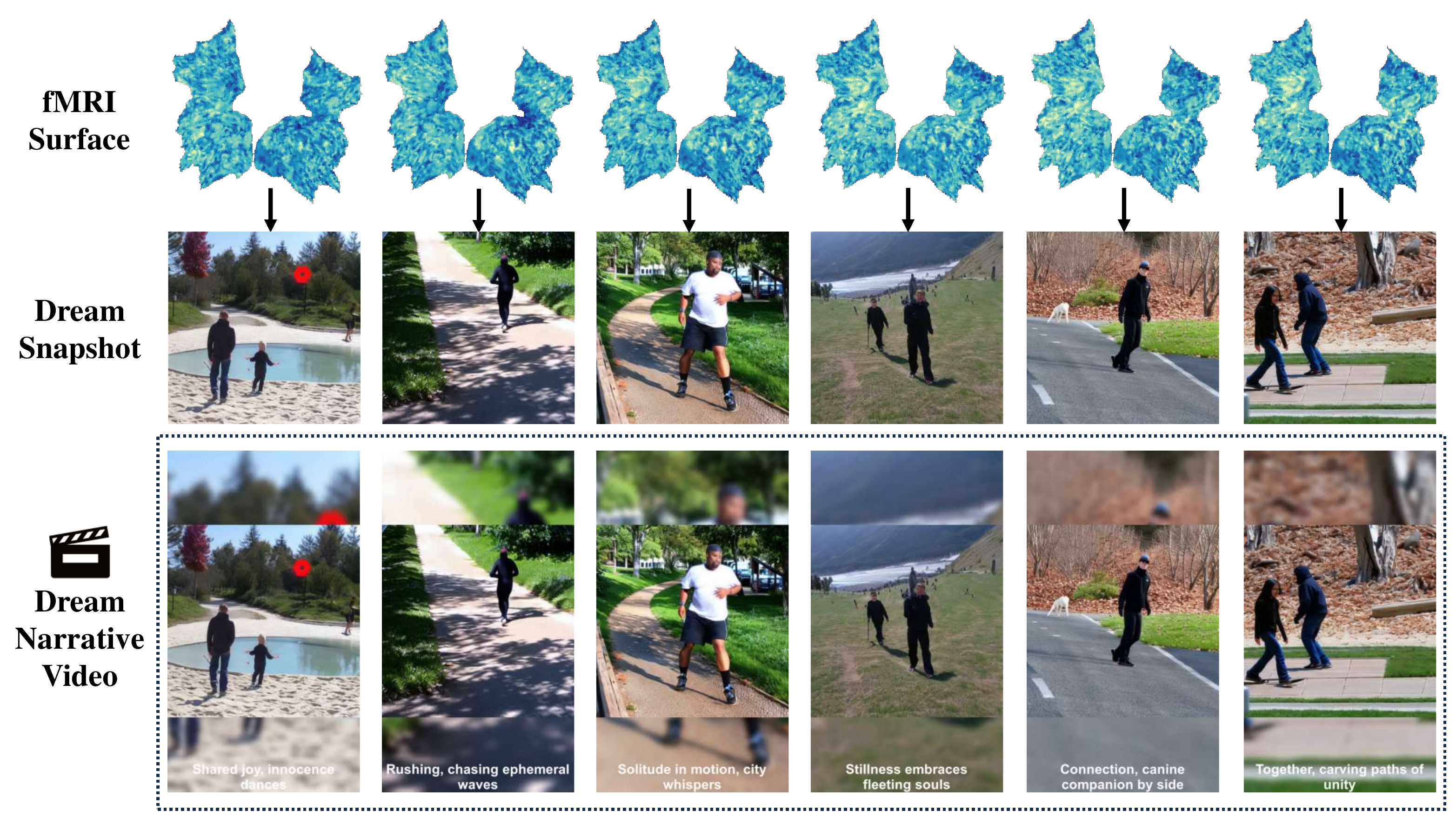}
  \caption{Visualization of Dream Narrative Video "people running". }
  \label{fig:run}
\end{figure*}

\subsection{Dataset}

% \noindent
% \textbf
\subsubsection{Natural Scenes Dataset.}
For the Visual Stimulus Perception Reconstruction task, we utilized all subjects from the Natural Scenes Dataset (NSD) \cite{allen2022massive} to train a multi-subject fMRI-to-image model. The NSD, collected using a 7T MRI scanner, is currently the largest-scale and highest-quality dataset for visual decoding. This enabled us to train the model using over 64,000 images and their corresponding fMRI signals from eight participants to reconstruct real visual perceptions.

% \noindent
% \textbf
\subsubsection{fMRI-Dream Dataset.}
The dream dataset we collected includes 3.3 hours of fMRI data from three participants during their sleep stages. Based on participants' dream reports, this consists of five instances of clearly recalled prolonged dream experiences and several brief, indistinct dreams that couldn't be precisely recalled. We selected the five instances with clear recall as ground truth labels to validate the accuracy of our dream interpretation. The other brief, indistinct dreams were excluded from consideration since participants couldn't definitively confirm their recollection. The five instances of clearly recalled dreams are detailed in Table \ref{tab:dream}.
We subsequently qualitatively and quantitatively analyze three of these segments. The performance of the other two samples is relatively lower due to unavoidable domain gaps and complex semantics. See supplementary materials for further analysis.

\begin{table}
  \caption{Instances of clearly recalled dreams. }
  \label{tab:freq}
  \renewcommand\arraystretch{1.2}
  \setlength{\tabcolsep}{20pt}
  \begin{tabular}{cl}
    \toprule
    Subject & Caption\\
    \midrule
    \multicolumn{1}{c}{\multirow{2}{*}{Sub-1}} & Skiing with a snowboard.\\
    & Enjoying a cup of milk tea.\\
    \midrule
    Sub-2 & People running.\\
    \midrule
    \multicolumn{1}{c}{\multirow{2}{*}{Sub-3}} & Some cats.\\
     & Some fruits, with plenty of grapes.\\
  \bottomrule
\end{tabular}
\label{tab:dream}
\vskip -0.1in
\end{table}

\subsection{Implementation Details}

During the Visual Stimulus Perception Reconstruction phase, we train the fMRI-to-image reconstruction model using data from 8 subjects from the NSD. In the Dream Visual Imagery Decoding and Dream Narrative Integration stages, we employ a single GTX 3090Ti GPU for image and text inference.

\subsection{Evaluation Metrics}
\label{sec:metric}

We utilize image category prediction to assess the alignment between reconstructed dream images and ground truth descriptions. Specifically, we augment the 80 class labels from the COCO dataset \cite{lin2014microsoft} with additional class labels extracted from the ground truth descriptions, merging identical labels to form the text categories used for dream evaluation. As these introduced labels differ from those in standard datasets, we employ CLIP \cite{radford2021learning} for zero-shot similarity computation. We organize text categories into captions using the template "a photo of [label]", then use CLIP to calculate the similarity between each reconstructed dream image and text, followed by softmax normalization to obtain the final image-category similarity.

\subsection{Qualitative Results}
We present visualizations of dream narratives for "cat," "skiing with a snowboard," and "people running" in Figures \ref{fig:cat}, \ref{fig:ski}, and \ref{fig:run}, respectively. It's evident that we initially decode individual dream images from the fMRI sequences, followed by assembling them into a cohesive dream narrative video. The resulting videos align with participants' descriptions. Leveraging the power of LLM, our constructed dream narratives are vivid, coherent, seamlessly weaving together disparate dream scenes. Complete videos are available in the supplementary materials.

\subsection{Quantitative Results}

To assess the correspondence between our decoded dreams and actual dream experiences, we partition the dataset into positive and negative samples for each genuine dream description in Table \ref{tab:dream}. Positive samples refer to the sleep fMRI data where participants reported the presence of the corresponding dream, while negative samples denote sleep fMRI data without that particular dream. Since the dreams of the three participants are distinct, for a given dream description, we can designate the corresponding participant's sleep data as positive samples and the sleep data from the other participants as negative samples.

\begin{figure}
  \centering
  \includegraphics[width=\linewidth]{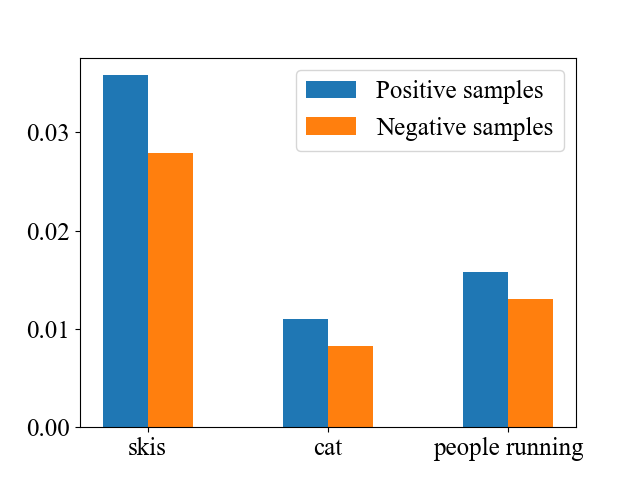}
  \vskip -0.1in
  \caption{Comparision of the average similarity between the ground-truth dream descriptions and the corresponding positive and negative samples' fMRI data. }
  \label{fig:sim}
  \vskip -0.1in
\end{figure}

Following the methodology outlined in Section \ref{sec:metric}, we compute the average similarity between the ground-truth dream descriptions and the corresponding positive and negative samples' fMRI data separately. The results are depicted in Figure \ref{fig:sim}. It can be observed that the average similarity of positive instances is consistently higher than that of negative instances, indicating a closer match between our decoded dream imagery and the actual dream descriptions.

Furthermore, we conducted standard Mann-Whitney U tests on the similarity sequences of positive and negative instances. As shown in Table \ref{tab:pvalue}, for the ground-truth dream descriptions "cat" and "people running," the p-value is smaller than $0.05$, indicating a significant difference in the distribution of positive and negative samples. For the description "skis," the p-value was $0.062$, suggesting a marginal difference in the distribution of positive and negative samples. While this result does not meet the conventional threshold of significance ($p < 0.05$), it still hints at a potential distinction.

\begin{table}
  \caption{P-value of Mann-Whitney U test for positive and negative samples. }
  \label{tab:freq}
  \renewcommand\arraystretch{1.2}
  \setlength{\tabcolsep}{10pt}
  \begin{tabular}{cccc}
  \toprule
    Dream Label & skis & cat & people running  \\ \midrule
        p-value & 0.062 & 0.0142 & 0.0001  \\
  \bottomrule
\end{tabular}
\label{tab:pvalue}
\end{table}

\section{Conclusions}
This paper has presented an innovative framework for converting dreams into coherent video narratives using fMRI data, marking a significant advancement in the field of multimedia and dream research. Our approach integrates cutting-edge techniques in fMRI analysis and language modeling to bridge the gap between subjective dream experiences and objective neurophysiological data. By reconstructing visual perceptions, decoding dream imagery, and integrating these elements into flowing narratives, we provide a novel method for visualizing and understanding dreams as comprehensive video stories.
The implications of our work extend beyond the scientific exploration of dreams. They open up new possibilities for creative expression, allowing individuals to explore their dream experiences. 

\bibliographystyle{ACM-Reference-Format}
\bibliography{sample-base}

\end{document}